\documentclass[10pt,conference,english]{IEEEtran}
\usepackage[utf8]{inputenc}
\usepackage[english]{babel}
\usepackage[T1]{fontenc}
\usepackage{amsfonts}
\usepackage{amsmath}
\usepackage{amssymb}
\usepackage{tikz}
\usepackage{graphicx}
\usepackage{color}
\usepackage{pgfpages}
\usepackage{enumitem}
\usetikzlibrary{arrows}
\usepackage{framed}
\usepackage{arydshln}
\usepackage{multirow}
\usepackage{url}
\usepackage{hyperref}
\usepackage{tablefootnote}

\title{Matching Convolutional Neural Networks\\without Priors about Data}

\author{
  \IEEEauthorblockN{Carlos Eduardo Rosar Kos Lassance\IEEEauthorrefmark{1}, Jean-Charles Vialatte\IEEEauthorrefmark{1}\IEEEauthorrefmark{2} and Vincent Gripon\IEEEauthorrefmark{1}}\\
\IEEEauthorblockA{\IEEEauthorrefmark{1}IMT Atlantique
    \\ \{carlos.rosarkoslassance, jc.vialatte, vincent.gripon\}@imt-atlantique.fr}
    \IEEEauthorblockA{\IEEEauthorrefmark{2}Cityzen Data
    }}

\newtheorem{definition}{Definition}

\begin{document}
\maketitle

\begin{abstract}
  We propose an extension of Convolutional Neural Networks (CNNs) to graph-structured data, including strided convolutions and data augmentation on graphs.
  Our method matches the accuracy of state-of-the-art CNNs when applied on images, without any prior about their 2D regular structure.
  On fMRI data, we obtain a significant gain in accuracy compared with existing graph-based alternatives.
\end{abstract}

\section{Introduction}

Convolutional Neural Networks (CNNs)~\cite{lecun1995convolutional} have been able to surpass traditional machine learning methods in various image based tasks \cite{he2016identity,krizhevsky2012imagenet}. 
This is possible as they exploit the learning capabilities of deep neural networks while also taking advantage of the intrinsic regular 2D structure of the data. But when data lacks regular structure~\cite{bronstein2017geometric}, there is no natural notion of convolutions, stride/pooling or data augmentation. Such irregularities occur in various domains covering social networks to neuroscience, internet of things, citation graphs, point cloud manifolds\dots The question of developing solutions that are counterparts of CNNs in irregular domains has recently been a very active field of research.

In this paper we introduce a method that extends CNNs to irregular domains. Contrary to many alternative works, we ensure that our proposed methodology matches the performance of CNNs when applied to regular domains, even without knowledge of the underlying structure. To that end, we infer a graph to represent the topology of the data. From this graph, we infer translations. The weight-sharing schemes of our proposed convolutional layers are then defined based on those translations, as well as data-augmentation and stride.

At the end of the process, the obtained architecture is very similar to a traditional CNN and can thus be trained using the same routines and libraries, and equivalent computational and memory footprints. We first perform experiments on the CIFAR-10 dataset without knowledge about the fact it is made of images. We show that our method is able to reach performance similar to state-of-art CNNs, thus implying that -- at least for regular domains -- it allows to completely leverage the underlying structure. Then, we perform experiments on an irregular neuroscience dataset and demonstrate a gain in performance compared with completely unstructured deep learning methods and alternative graph-based CNNs.


\section{Related Work}
\label{related_works}
Deep learning on graphs can refer to three distinct problems: classification of graphs, of nodes in a graph, or of signals on graphs. In this paper, we are interested only in the latter task that is to leverage the graph structure of signals in deep learning models, by redefining the convolutional layer. Such methods have already been proposed in the literature. We distinguish two categories of solutions.

In the first category, convolution is defined as pointwise multiplication in the spectral domain of the graph, which is defined using the Laplace-Beltrami operator~\cite{chung1996spectral}. This method originated the first spectral graph CNNs~\cite{bruna2013spectral,henaff2015deep}. An approximation of the spectral graph convolution using Chebychev polynomials has been proposed~\cite{defferrard2016convolutional}, and has the advantage to be both faster and localized in the vertex domain. Another variant with Cayley polynomials~\cite{levie2017cayleynets} also localizes the convoluted filter in the spectral domain.

In the second category, convolution is defined directly in the vertex domain. These works were originally motivated by chemistry datasets~\cite{duvenaud2015convolutional,kearnes2016molecular}. Convolution is defined as a function of the kernel weights and neighboring vertices (the receptive field), usually based on dot products. As such, it retains the property of being localized and of sharing weights. But there remains the need to specify how the shared weights are allocated in this receptive field~\cite{vialatte2016generalizing}. This allocation can depend on an arbitrary order~\cite{niepert2016learning}, on the number of hops~\cite{atwood2016diffusion,du2017topology}, on both vertices and their neighbors~\cite{monti2016geometric,simonovsky2017dynamic}, on another learned kernel~\cite{vialatte2017learning}, on an attention mechanism~\cite{velickovic2017graph}, on pattern identification~\cite{sankar2017motif}, or on translation identification~\cite{pasdeloup2017convolutional}. All these methods also differ in the function that maps the receptive fields and the weight kernel to the neuron's outputs. But in the end, these definitions overlap. That is why some authors have proposed unified frameworks~\cite{gilmer2017neural}.


We tackle another point. Given a dataset with no structure between the features of the input vectors, our goal is to demonstrate that we can still define meaningful convolutional, stride/pooling layers. The first step to determine whether the results obtained on unstructured data is satisfactory or not is to stress it on regular data while disregarding its structure. We deal with this step on image datasets. Even though some previous works match the performance of CNNs on image datasets, ours is the first that can do it without structure prior.

\section{Methodology}
\label{methodology}

\begin{figure}[h]
\begin{framed}
  Step 0 (optional): infer a graph
  \begin{center}
    \begin{tikzpicture}[thick]
      \node at (0,0) {$\mathbf{x}_0$};
      \path[black!20!white, dashed]
      (0.5,0) edge (2.5,0);
      \path
      (0.75, 0) edge[blue] (0.75, 0.13)
      (1.25, 0) edge[blue] (1.25, 0.1)
      (1.75, 0) edge[red] (1.75, -0.1)
      (2.25, 0) edge[red] (2.25, -0.05);
      \node[fill, circle, inner sep = 1pt, blue] at (0.75, 0.13) {};
      \node[fill, circle, inner sep = 1pt, blue] at (1.25, 0.1) {};
      \node[fill, circle, inner sep = 1pt, red] at (1.75, -0.1) {};
      \node[fill, circle, inner sep = 1pt, red] at (2.25, -0.05) {};
      \node at (0,-0.25) {$\mathbf{x}_1$};
      \begin{scope}[yshift=-0.25cm]
        \path
        (0.75, 0) edge[blue] (0.75, 0.07)
        (1.25, 0) edge[blue] (1.25, 0.08)
        (1.75, 0) edge[red] (1.75, -0.08)
        (2.25, 0) edge[red] (2.25, -0.15);
        \node[fill, circle, inner sep = 1pt,blue] at (0.75, 0.07){};
        \node[fill, circle, inner sep = 1pt,blue] at (1.25, 0.08){};
        \node[fill, circle, inner sep = 1pt,red] at (1.75, -0.08){};
        \node[fill, circle, inner sep = 1pt,red] at (2.25, -0.15){};        
      \end{scope}

      \path[black!20!white, dashed]
      (0.5,-0.25) edge (2.5,-0.25);
      \node at (0,-0.5) {\vdots};
      \node at (0,-1) {$\mathbf{x}_m$};
      \begin{scope}[yshift=-1cm]
        \path
        (0.75, 0) edge[blue] (0.75, 0.12)
        (1.25, 0) edge[red] (1.25, -0.05)
        (1.75, 0) edge[red] (1.75, -0.03)
        (2.25, 0) edge[blue] (2.25, 0.08);
        \node[fill, circle, inner sep = 1pt, blue] at (0.75, 0.12){};
        \node[fill, circle, inner sep = 1pt, red] at (1.25, -0.05){};
        \node[fill, circle, inner sep = 1pt, red] at (1.75, -0.03){};
        \node[fill, circle, inner sep = 1pt, blue] at (2.25, 0.08){};        
      \end{scope}
      \node at (1.5,-0.5) {\vdots};
      \path[black!20!white, dashed]
      (0.5,-1) edge (2.5,-1);

      \node at (3.25, -0.5) {\huge{$\Rightarrow$}};

      \begin{scope}[xshift=4cm]
        \node[draw, circle](a) at (0,0) {1};
        \node[draw, circle](b) at (1.5,0) {2};
        \node[draw, circle](c) at (1.5,-1) {4};
        \node[draw, circle](d) at (0,-1) {3};
      \end{scope}
      \path
      (a) edge (b)
      edge (d)
      (b) edge (d)
      (c) edge (d);
    \end{tikzpicture}
  \end{center}
  
  Step 1: infer translations (Subsection B)
  \begin{center}
    \begin{tikzpicture}[thick]      
      \begin{scope}[xshift=0cm]
        \node[draw, circle](a) at (0,0) {1};
        \node[draw, circle](b) at (1.5,0) {2};
        \node[draw, circle](c) at (1.5,-1) {4};
        \node[draw, circle](d) at (0,-1) {3};
      \end{scope}
      \path
      (a) edge (b)
      edge (d)
      (b) edge (d)
      (c) edge (d);
      \node at (2.75, -0.5) {\huge{$\Rightarrow$}};
      \begin{scope}[xshift=4cm, scale=0.5, yshift=0.5cm]
        \node[draw, inner sep=2pt, circle](a) at (0,0) {};
        \node[draw, inner sep=2pt, circle](b) at (1.5,0) {};
        \node[draw, inner sep=2pt, circle](c) at (1.5,-1) {};
        \node[draw, inner sep=2pt, circle](d) at (0,-1) {};
      \end{scope}
      \path[black!20!white]
      (a) edge (b)
      edge (d)
      (b) edge (d)
      (c) edge (d);
      \path[->,>=stealth']
      (a) edge (b)      
      (d) edge (c);
      \begin{scope}[xshift=4cm, scale=0.5, yshift=-1.5cm]
        \node[draw, inner sep=2pt, circle](a) at (0,0) {};
        \node[draw, inner sep=2pt, circle](b) at (1.5,0) {};
        \node[draw, inner sep=2pt, circle](c) at (1.5,-1) {};
        \node[draw, inner sep=2pt, circle](d) at (0,-1) {};
      \end{scope}
      \path[black!20!white]
      (a) edge (b)
      edge (d)
      (b) edge (d)
      (c) edge (d);
      \path[->,>=stealth']
      (b) edge (a)      
      (c) edge (d);
      \begin{scope}[xshift=5.5cm, scale=0.5, yshift=0.5cm]
        \node[draw, inner sep=2pt, circle](a) at (0,0) {};
        \node[draw, inner sep=2pt, circle](b) at (1.5,0) {};
        \node[draw, inner sep=2pt, circle](c) at (1.5,-1) {};
        \node[draw, inner sep=2pt, circle](d) at (0,-1) {};
      \end{scope}
      \path[black!20!white]
      (a) edge (b)
      edge (d)
      (b) edge (d)
      (c) edge (d);
      \path[->,>=stealth']
      (a) edge (d);
      \begin{scope}[xshift=5.5cm, scale=0.5, yshift=-1.5cm]
        \node[draw, inner sep=2pt, circle](a) at (0,0) {};
        \node[draw, inner sep=2pt, circle](b) at (1.5,0) {};
        \node[draw, inner sep=2pt, circle](c) at (1.5,-1) {};
        \node[draw, inner sep=2pt, circle](d) at (0,-1) {};
      \end{scope}
      \path[black!20!white]
      (a) edge (b)
      edge (d)
      (b) edge (d)
      (c) edge (d);
      \path[->,>=stealth']
      (d) edge (a);
    \end{tikzpicture}
  
  \end{center}
  
  Step 2: design convolution weight-sharing (Subsection C)

  \begin{center}
    \begin{tikzpicture}[thick]      
      \begin{scope}[xshift=0cm, scale=0.75, yshift=0.5cm]
        \node[inner sep=2pt](w) at (-1.0,-0.5) {$w_1 \times$};
        \node[draw, inner sep=2pt, circle](a) at (0,0) {};
        \node[draw, inner sep=2pt, circle](b) at (1.5,0) {};
        \node[draw, inner sep=2pt, circle](c) at (1.5,-1) {};
        \node[draw, inner sep=2pt, circle](d) at (0,-1) {};       
      \end{scope}
      \path[black!20!white]
      (a) edge (b)
      edge (d)
      (b) edge (d)
      (c) edge (d);
      \path[->,>=stealth']
      (a) edge (b)      
      (d) edge (c);
      \begin{scope}[xshift=0cm, scale=0.75, yshift=-1.5cm]
        \node[inner sep=2pt](w) at (-1.0,-0.5) {+ $w_2 \times$};
        \node[draw, inner sep=2pt, circle](a) at (0,0) {};
        \node[draw, inner sep=2pt, circle](b) at (1.5,0) {};
        \node[draw, inner sep=2pt, circle](c) at (1.5,-1) {};
        \node[draw, inner sep=2pt, circle](d) at (0,-1) {};
      \end{scope}
      \path[black!20!white]
      (a) edge (b)
      edge (d)
      (b) edge (d)
      (c) edge (d);
      \path[->,>=stealth']
      (b) edge (a)      
      (c) edge (d);
      \begin{scope}[xshift=2.5cm, scale=0.75, yshift=0.5cm]
        \node[inner sep=2pt](w) at (-1.0,-0.5) {+ $w_3 \times$};
        \node[draw, inner sep=2pt, circle](a) at (0,0) {};
        \node[draw, inner sep=2pt, circle](b) at (1.5,0) {};
        \node[draw, inner sep=2pt, circle](c) at (1.5,-1) {};
        \node[draw, inner sep=2pt, circle](d) at (0,-1) {};
      \end{scope}
      \path[black!20!white]
      (a) edge (b)
      edge (d)
      (b) edge (d)
      (c) edge (d);
      \path[->,>=stealth']
      (a) edge (d);
      \begin{scope}[xshift=2.5cm, scale=0.75, yshift=-1.5cm]
        \node[inner sep=2pt](w) at (-1.0,-0.5) {+ $w_4 \times$};      
        \node[draw, inner sep=2pt, circle](a) at (0,0) {};
        \node[draw, inner sep=2pt, circle](b) at (1.5,0) {};
        \node[draw, inner sep=2pt, circle](c) at (1.5,-1) {};
        \node[draw, inner sep=2pt, circle](d) at (0,-1) {};
      \end{scope}
      \path[black!20!white]
      (a) edge (b)
      edge (d)
      (b) edge (d)
      (c) edge (d);
      \path[->,>=stealth']
      (d) edge (a);

      \begin{scope}[xshift=5.0cm, scale=0.75, yshift=-0.5cm]
        \node[inner sep=2pt](w) at (-1.0,-0.5) {+ $w_0 \times$};
        \node[draw, inner sep=2pt, circle](a) at (0,0) {};
        \node[draw, inner sep=2pt, circle](b) at (1.5,0) {};
        \node[draw, inner sep=2pt, circle](c) at (1.5,-1) {};
        \node[draw, inner sep=2pt, circle](d) at (0,-1) {};
      \end{scope}
      \path[black!20!white]
      (a) edge (b)
      edge (d)
      (b) edge (d)
      (c) edge (d);
      \path[]
      (a) edge [loop above] (a)
      (b) edge [loop above] (b)
      (c) edge [loop below] (c)
      (d) edge [loop below] (d);

    \end{tikzpicture}
  
  \end{center}
  
  Step 3: design data-augmentation (Subsection D)
  \begin{center}
    \begin{tikzpicture}[thick]      
      \begin{scope}[xshift=0cm]
        \node[draw, inner sep = 2pt, circle](a) at (0,0) {};
        \node[draw, inner sep = 2pt, circle](b) at (1.5,0) {};
        \node[draw, inner sep = 2pt, circle](c) at (1.5,-1) {};
        \node[draw, inner sep = 2pt, circle](d) at (0,-1) {};
        \path[]
        (0, 0) edge[blue] (0, 0.52)
        (1.5, 0) edge[blue] (1.5, 0.4)
        (0, -1) edge[red] (0, -1.4)
        (1.5, -1) edge[red] (1.5, -1.2);
        \node[fill, circle, inner sep = 1pt, blue] at (0, 0.52) {};
        \node[fill, circle, inner sep = 1pt, blue] at (1.5, 0.4) {};
        \node[fill, circle, inner sep = 1pt, red] at (0, -1.4) {};
        \node[fill, circle, inner sep = 1pt, red] at (1.5, -1.2) {};
      \end{scope}
      \node at (-0.5,-0.5) {$\mathbf{x}_0$};
      \path[black!20!white]
      (a) edge (b)
      edge (d)
      (b) edge (d)
      (c) edge (d);
      \node at (2.25, -0.5) {\huge{$\Rightarrow$}};
      \begin{scope}[xshift=3cm, scale=0.5, yshift=0.75cm]
        \node[draw, inner sep=2pt, circle](a) at (0,0) {};
        \node[draw, inner sep=2pt, circle](b) at (1.5,0) {};
        \node[draw, inner sep=2pt, circle](c) at (1.5,-1) {};
        \node[draw, inner sep=2pt, circle](d) at (0,-1) {};
      \end{scope}
      \path[black!20!white]
      (a) edge (b)
      edge (d)
      (b) edge (d)
      (c) edge (d);
      \path[->,>=stealth']
      (a) edge (b)      
      (d) edge (c);
      \begin{scope}[xshift=4.5cm, scale=0.5, yshift = 0.75cm]
        \node[draw, inner sep = 2pt, circle](a) at (0,0) {};
        \node[draw, inner sep = 2pt, circle](b) at (1.5,0) {};
        \node[draw, inner sep = 2pt, circle](c) at (1.5,-1) {};
        \node[draw, inner sep = 2pt, circle](d) at (0,-1) {};
        \path[]
        (1.5, 0) edge[blue] (1.5, 0.52)
        (1.5, -1) edge[red] (1.5, -1.4);
        \node[fill, circle, inner sep = 1pt, blue] at (1.5, 0.52) {};
        \node[fill, circle, inner sep = 1pt, red] at (1.5, -1.4) {};
        \path[black!20!white]
      (a) edge (b)
      edge (d)
      (b) edge (d)
      (c) edge (d);
      \end{scope}      
      \begin{scope}[xshift=3cm, scale=0.5, yshift=-1.75cm]
        \node[draw, inner sep=2pt, circle](a) at (0,0) {};
        \node[draw, inner sep=2pt, circle](b) at (1.5,0) {};
        \node[draw, inner sep=2pt, circle](c) at (1.5,-1) {};
        \node[draw, inner sep=2pt, circle](d) at (0,-1) {};
      \end{scope}
      \path[black!20!white]
      (a) edge (b)
      edge (d)
      (b) edge (d)
      (c) edge (d);
      \path[->,>=stealth']
      (b) edge (a)      
      (c) edge (d);
      \begin{scope}[xshift=4.5cm, scale=0.5, yshift=-1.75cm]
        \node[draw, inner sep = 2pt, circle](a) at (0,0) {};
        \node[draw, inner sep = 2pt, circle](b) at (1.5,0) {};
        \node[draw, inner sep = 2pt, circle](c) at (1.5,-1) {};
        \node[draw, inner sep = 2pt, circle](d) at (0,-1) {};
        \path[]
        (0, 0) edge[blue] (0, 0.4)
        (0, -1) edge[red] (0, -1.2);
        \node[fill, circle, inner sep = 1pt, blue] at (0, 0.4) {};
        \node[fill, circle, inner sep = 1pt, red] at (0, -1.2) {};
        \path[black!20!white]
      (a) edge (b)
      edge (d)
      (b) edge (d)
      (c) edge (d);
      \end{scope}
    \end{tikzpicture}
  
  \end{center}
  Step 4: design graph subsampling and convolution weight-sharing (Subsection E)

  \begin{center}
    \begin{tikzpicture}[thick, scale=0.7]
      \tikzstyle{every node} = [inner sep = 3pt];
      \small{\begin{scope}[xshift=-0.5cm]
        \node[draw, circle](a) at (0,0) {1};
        \node[draw, circle](b) at (1.5,0) {2};
        \node[draw, circle](c) at (1.5,-1) {4};
        \node[draw, circle](d) at (0,-1) {3};
      \end{scope}
      \path
      (a) edge (b)
      edge (d)
      (b) edge (d)
      (c) edge (d);
      \node at (2, -0.5) {\huge{$\Rightarrow$}};
      \begin{scope}[xshift=3.25cm]
        \node[draw, circle](a) at (0,0) {1};
        \node[draw, circle, black!20!white](b) at (1.5,0) {2};
        \node[draw, circle](c) at (1.5,-1) {4};
        \node[draw, circle, black!20!white](d) at (0,-1) {3};
      \end{scope}
      \path
      (a) edge (c);
      \node at (6, -0.5) {\huge{$\Rightarrow$}};
      \begin{scope}[xshift=7.75cm, scale=0.4, yshift=0.5cm]
        \node[inner sep=1pt](w) at (-2.0,-0.5) {$w_0 \times$};
        \node[draw, inner sep=1.5pt, circle](a) at (0,0) {};
        \node[draw, inner sep=1.5pt, circle, black!20!white](b) at (1.5,0) {};
        \node[draw, inner sep=1.5pt, circle](c) at (1.5,-1) {};
        \node[draw, inner sep=1.5pt, circle, black!20!white](d) at (0,-1) {};
      \end{scope}
      \path[black!20!white]
      (a) edge (c);
      \path[]
      (a) edge [loop above] (a)
      (c) edge [loop below] (c);
      \begin{scope}[xshift=10cm, scale=0.4, yshift=0.5cm]
        \node[inner sep=1pt](w) at (-2.0,-0.5) {+ $w_1 \times$};
        \node[draw, inner sep=1.5pt, circle](a) at (0,0) {};
        \node[draw, inner sep=1.5pt, circle, black!20!white](b) at (1.5,0) {};
        \node[draw, inner sep=1.5pt, circle](c) at (1.5,-1) {};
        \node[draw, inner sep=1.5pt, circle, black!20!white](d) at (0,-1) {};
      \end{scope}
      \path[black!20!white]
      (a) edge (c);
      \path[]
      (c) edge [->,>=stealth'] (a);

      \begin{scope}[xshift=7.75cm, scale=0.4, yshift=-2.5cm]
        \node[inner sep=1pt](w) at (-2.0,-0.5) {+ $w_2 \times$};
        \node[draw, inner sep=1.5pt, circle](a) at (0,0) {};
        \node[draw, inner sep=1.5pt, circle, black!20!white](b) at (1.5,0) {};
        \node[draw, inner sep=1.5pt, circle](c) at (1.5,-1) {};
        \node[draw, inner sep=1.5pt, circle, black!20!white](d) at (0,-1) {};
      \end{scope}
      \path[black!20!white]
      (a) edge (c);
      \path[->,>=stealth']
      (a) edge [] (c);
	
      }
    \end{tikzpicture}
  
  \end{center}
	\end{framed}
  \caption{Outline of the proposed method}
  \label{outline}
  \vspace{-.5cm}
\end{figure}
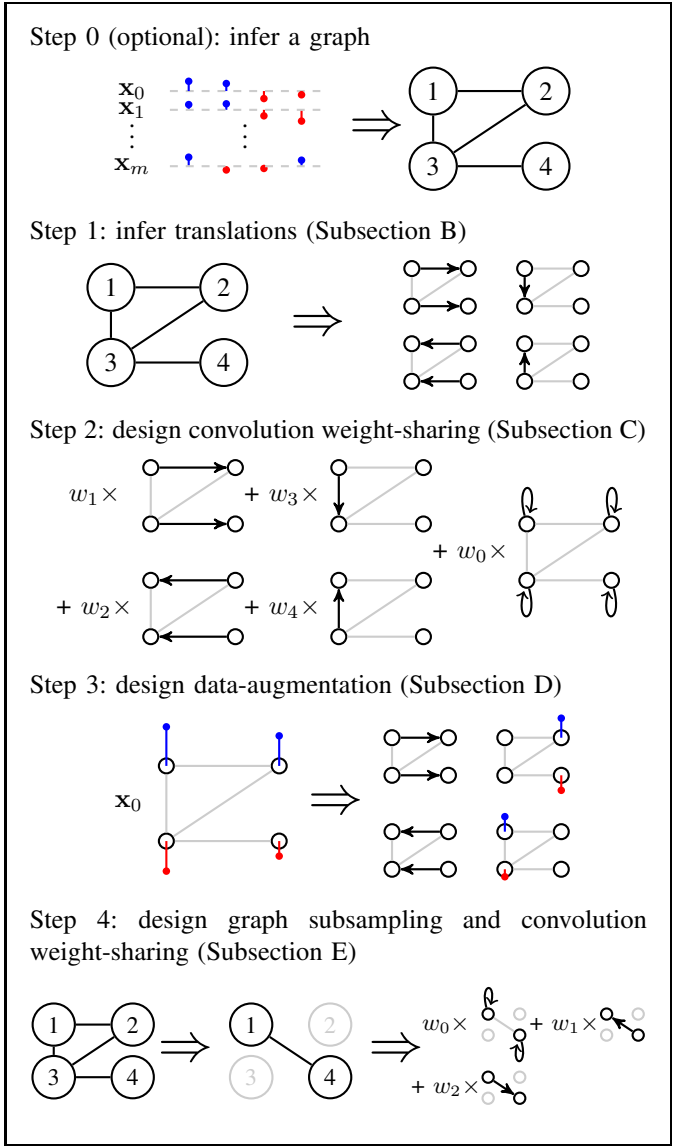

Our method is based on~\cite{pasdeloup2017convolutional}, where the authors have introduced a way to infer a graph from training signals, then translations from the obtained graph to design ad-hoc CNNs. We extend this approach and design strided convolutions along graph downscaling, data augmentation and convolutions on downscaled graphs. Figure~\ref{outline} depicts the proposed method.

\subsection{Background}


Define a graph $G = \langle V, E \rangle$ with $V$ the set of vertices, and $E \subseteq\binom{V}{2}$ the set of edges. We suppose the graph is connected, as conversely the process can be applied to each connected component of $G$. We denote by $d$ the max degree of the graph and $n = |V|$ the number of vertices.

The authors of~\cite{pasdeloup2017convolutional} propose to inductively define translations as functions from vertices to vertices as follows:

\begin{definition}{\textbf{Candidate-translation}}\\
  A \emph{candidate-translation} is a function $\phi: U \to V$, where $U \subset V$ and such that:
  \begin{itemize}[noitemsep,nolistsep]
  \item $\phi$ is \emph{injective}:\\
  $\forall v,v' \in U, \phi(v) = \phi(v') \Rightarrow v = v',$
  \item $\phi$ is \emph{edge-constrained}:\\
  $\forall v \in U, (v,\phi(v)) \in E,$
  \item $\phi$ is \emph{strongly neighborhood-preserving}:\\
  $\forall v,v' \in U, (v,v')\in E \Leftrightarrow (\phi(v),\phi(v')) \in E.$\\
  \end{itemize}
\end{definition}

The cardinal $|V-U|$ is called the \emph{loss} of $\phi$.
Two candidate-translations $\phi$ and $\phi'$ are said to be \emph{aligned} if $\exists v\in V, \phi(v) = \phi'(v)$.
We define $N_r(v)$ as the set of vertices that are at most $r$-hop away from a vertex $v \in V$.\\

\begin{definition}{\textbf{Translation}}\\
  A \emph{translation} in a graph $G$ is a candidate-translation such that there is no aligned translation with a strictly smaller loss, or is the identity function.\\
\end{definition}

Note that if the graph is a 2D grid, obtained translations are exactly natural translations on images~\cite{GrePasViaGri201610}.\\

\begin{definition}{\textbf{Local translation}}\\
A \emph{local translation} of center $v \in V$ is a translation in the subgraph of $G$ induced by $N_2(v)$, that has $v$ in its definition domain.\\
\end{definition}

As local translations can't be used to design data augmentation and convolutions on downscaled graphs, we also design proxies to global translations.\\

\begin{definition}{\textbf{Proxy-translations}}\\
A family of \emph{proxy-translations} $(\psi_p)_{p=0,..\kappa-1}$ initialized by $v_0 \in V$ is defined algorithmically as follows:
\begin{enumerate}
\item We place an indexing kernel on $N_1(v_0) $ i.e.\\ $N_1(v_0) = \{v_0, v_1, ..., v_{\kappa-1}\}$ with $\forall p, \psi_p(v_0) = v_p$,
\item We move this kernel using each local translation $\phi$ of center $v_0$: $\forall p, \psi_p(\phi(v_0)) = \phi(v_p)$,
\item We repeat 2) from each new center reached until saturation. If a center is being reached again, we keep the indexing that minimizes the sum of losses of the local translations that has lead to it.
\end{enumerate}
\end{definition}

\subsection{Efficiently Finding Translations}
\label{trans}

Finding translations is an NP-complete problem~\cite{pasdeloup2017translations}, such that for large graphs the method is not suitable. 
In order to break down complexity, the authors of~\cite{pasdeloup2017convolutional} propose to search for local translations. They also introduce approximate translations which we omit for the sake of simplicity, but the description would be similar. We describe in three steps how we efficiently find proxy-translations.\\

\noindent\textbf{First step: finding local translations}

For each vertex $v \in G$, we identify all local translations using a bruteforce algorithm. This process requires finding all translations in all induced subgraphs. There are $n$ such subgraphs, each one contains at most $d$ local translations. Finding a translation can be performed by looking at all possible injections from 1-hop vertices around the central vertex to any vertex that is at most 2-hops away. We conclude that it requires at most $\mathcal{O}(nd d^{2(d+1)})$ elementary operations and is thus linear with the order of the graph. On the other hand, it suggests that sparsity of the graph is a key criterion in order to maintain the complexity reasonable.

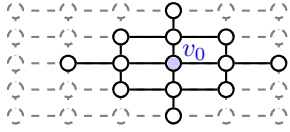
\begin{figure}
  \begin{center}
    \begin{tikzpicture}[thick, scale = 0.7]
      \tikzstyle{every node} = [inner sep = 2pt];
      \foreach \x in {0,...,5}{
        \foreach \y in {0,...,4}{
          \node [draw, circle, dashed, black!50] at (\x, \y/2) {};
        }
      }
      \foreach \x in {0,...,4}{
        \foreach \y in {0,...,4}{
          \path[-]
          (\x,\y/2) edge[dashed, black!50] (\x+1,\y/2);
        }
      }
      \foreach \x in {0,...,5}{
        \foreach \y in {0,...,3}{
          \path[-]
          (\x,\y/2) edge[dashed, black!50] (\x,\y/2 + 1/2);
        }
      }
      \foreach \x in {0,...,5}{
        \foreach \y in {0,...,4}{
          \node [draw = black!50, dashed, circle,fill=white] at (\x, \y/2) {};
        }
      }
      \node[fill=blue!20!white] at (3,1){};
      \node at (3.4,1.2){\textcolor{blue}{$v_0$}};
      \node(bb) [draw, circle] at (3, 0) {};
      \node(s) [draw, circle] at (3, 1) {};
      \node(b) [draw, circle] at (3, 0.5) {};
      \node(u) [draw, circle] at (3, 1.5) {};
      \node(uu) [draw, circle] at (3, 2) {};
      \node(l) [draw, circle] at (2, 1) {};
      \node(lb) [draw, circle] at (2, 0.5) {};
      \node(lu) [draw, circle] at (2, 1.5) {};
      \node(r) [draw, circle] at (4, 1) {};
      \node(rb) [draw, circle] at (4, 0.5) {};
      \node(ru) [draw, circle] at (4, 1.5) {};
      \node(ll) [draw, circle] at (1, 1) {};
      \node(rr) [draw, circle] at (5, 1) {};
      \path[]
      (s) edge (u)
      edge (l)
      edge (b)
      edge (r)
      (l) edge (ll)
      edge (lu)
      edge (lb)
      (r) edge (rr)
      edge (ru)
      edge (rb)
      (u) edge (uu)
      edge (lu)
      edge (ru)
      (b) edge (bb)
      edge (lb)
      edge (rb);
    \end{tikzpicture}
  \end{center}
  \caption{Grid graph (in dashed grey) and the subgraph induced by $N_2(v_0)$ (in black).}
  \label{gridgraph}
\end{figure}

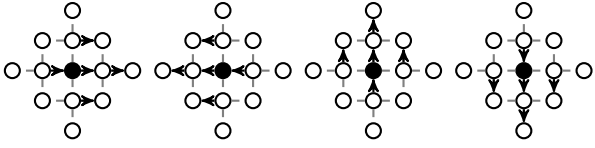
\begin{figure}
  \begin{center}
    \begin{tikzpicture}[thick]
      \begin{scope}[scale=0.8]
        \tikzstyle{every node} = [inner sep=2pt]
        \node[fill=black, circle] at (3,1){};
        \node(bb) [draw, circle] at (3, 0) {};
        \node(s) [draw, circle] at (3, 1) {};
        \node(b) [draw, circle] at (3, 0.5) {};
        \node(u) [draw, circle] at (3, 1.5) {};
        \node(uu) [draw, circle] at (3, 2) {};
        \node(l) [draw, circle] at (2.5, 1) {};
        \node(lb) [draw, circle] at (2.5, 0.5) {};
        \node(lu) [draw, circle] at (2.5, 1.5) {};
        \node(r) [draw, circle] at (3.5, 1) {};
        \node(rb) [draw, circle] at (3.5, 0.5) {};
        \node(ru) [draw, circle] at (3.5, 1.5) {};
        \node(ll) [draw, circle] at (2, 1) {};
        \node(rr) [draw, circle] at (4, 1) {};
        \path[black!50, dashed]
        (s) edge (u)
        edge (l)
        edge (b)
        edge (r)
        (l) edge (ll)
        edge (lu)
        edge (lb)
        (r) edge (rr)
        edge (ru)
        edge (rb)
        (u) edge (uu)
        edge (lu)
        edge (ru)
        (b) edge (bb)
        edge (lb)
        edge (rb);
        \path[->,>=stealth']
        (s) edge (r)
        (r) edge (rr)
        (u) edge (ru)
        (b) edge (rb)
        (l) edge (s)
        ;
      \end{scope}
      \begin{scope}[xshift=2cm,scale=0.8]
        \tikzstyle{every node} = [inner sep=2pt]
      \node[fill=black, circle] at (3,1){};
      \node(bb) [draw, circle] at (3, 0) {};
      \node(s) [draw, circle] at (3, 1) {};
      \node(b) [draw, circle] at (3, 0.5) {};
      \node(u) [draw, circle] at (3, 1.5) {};
      \node(uu) [draw, circle] at (3, 2) {};
      \node(l) [draw, circle] at (2.5, 1) {};
      \node(lb) [draw, circle] at (2.5, 0.5) {};
      \node(lu) [draw, circle] at (2.5, 1.5) {};
      \node(r) [draw, circle] at (3.5, 1) {};
      \node(rb) [draw, circle] at (3.5, 0.5) {};
      \node(ru) [draw, circle] at (3.5, 1.5) {};
      \node(ll) [draw, circle] at (2, 1) {};
      \node(rr) [draw, circle] at (4, 1) {};
      \path[black!50, dashed]
      (s) edge (u)
      edge (l)
      edge (b)
      edge (r)
      (l) edge (ll)
      edge (lu)
      edge (lb)
      (r) edge (rr)
      edge (ru)
      edge (rb)
      (u) edge (uu)
      edge (lu)
      edge (ru)
      (b) edge (bb)
      edge (lb)
      edge (rb);
      \path[->,>=stealth']
      (s) edge (l)
      (r) edge (s)
      (u) edge (lu)
      (b) edge (lb)
      (l) edge (ll)
      ;
      \end{scope}
      \begin{scope}[xshift=4cm, scale=0.8]
        \tikzstyle{every node} = [inner sep=2pt]
        \node[fill=black, circle] at (3,1){};
        \node(bb) [draw, circle] at (3, 0) {};
        \node(s) [draw, circle] at (3, 1) {};
        \node(b) [draw, circle] at (3, 0.5) {};
        \node(u) [draw, circle] at (3, 1.5) {};
        \node(uu) [draw, circle] at (3, 2) {};
        \node(l) [draw, circle] at (2.5, 1) {};
        \node(lb) [draw, circle] at (2.5, 0.5) {};
        \node(lu) [draw, circle] at (2.5, 1.5) {};
        \node(r) [draw, circle] at (3.5, 1) {};
        \node(rb) [draw, circle] at (3.5, 0.5) {};
        \node(ru) [draw, circle] at (3.5, 1.5) {};
        \node(ll) [draw, circle] at (2, 1) {};
        \node(rr) [draw, circle] at (4, 1) {};
        \path[black!50, dashed]
        (s) edge (u)
        edge (l)
        edge (b)
        edge (r)
        (l) edge (ll)
        edge (lu)
        edge (lb)
        (r) edge (rr)
        edge (ru)
        edge (rb)
        (u) edge (uu)
        edge (lu)
        edge (ru)
        (b) edge (bb)
        edge (lb)
        edge (rb);
        \path[->,>=stealth']
        (s) edge (u)
        (r) edge (ru)
        (u) edge (uu)
        (b) edge (s)
        (l) edge (lu)
        ;
      \end{scope}
      \begin{scope}[xshift=6cm, scale=0.8]
        \tikzstyle{every node} = [inner sep=2pt]
        \node[fill=black, circle] at (3,1){};
        \node(bb) [draw, circle] at (3, 0) {};
        \node(s) [draw, circle] at (3, 1) {};
        \node(b) [draw, circle] at (3, 0.5) {};
        \node(u) [draw, circle] at (3, 1.5) {};
        \node(uu) [draw, circle] at (3, 2) {};
        \node(l) [draw, circle] at (2.5, 1) {};
        \node(lb) [draw, circle] at (2.5, 0.5) {};
        \node(lu) [draw, circle] at (2.5, 1.5) {};
        \node(r) [draw, circle] at (3.5, 1) {};
        \node(rb) [draw, circle] at (3.5, 0.5) {};
        \node(ru) [draw, circle] at (3.5, 1.5) {};
        \node(ll) [draw, circle] at (2, 1) {};
        \node(rr) [draw, circle] at (4, 1) {};
        \path[black!50, dashed]
        (s) edge (u)
        edge (l)
        edge (b)
        edge (r)
        (l) edge (ll)
        edge (lu)
        edge (lb)
        (r) edge (rr)
        edge (ru)
        edge (rb)
        (u) edge (uu)
        edge (lu)
        edge (ru)
        (b) edge (bb)
        edge (lb)
        edge (rb);
        \path[->,>=stealth']
        (s) edge (b)
        (r) edge (rb)
        (u) edge (s)
        (b) edge (bb)
        (l) edge (lb)
        ;
      \end{scope}
    \end{tikzpicture}
  \end{center}
  \caption{Translations (black arrows) in the induced subgraph (dashed grey) around $v_0$ (filled in black) that contains $v_0$ and only some of its neighbors.}
  \label{inducedtranslations}
\end{figure}

Figure~\ref{gridgraph} depicts an example of a grid graph and the induced subgraph around vertex $v_0$. Figure~\ref{inducedtranslations} depicts all obtained translations in the induced subgraph.\\

\noindent\textbf{Second step: using local translations to move a small localized kernel around $G$}

Given an arbitrary\footnote{In practice we run several experiments while changing the initial vertex and keep the best obtained result.} vertex $v_0 \in V$, we place an indexing kernel on $N_1(v_0)$ i.e. $N_1(v_0) = \{v_0, v_1, ..., v_{\kappa-1}\}$. Then we move it using every local translations of center $v_0$, repeating this process for each center that is reached for the first time. We stop when the kernel has been moved everywhere in the graph. In case of multiple paths leading to the same destination, we keep the indexing that minimizes the sum of loss of the series of local translations. We henceforth obtain an indexing of at most $\kappa$ objects of $N_1(v)$ for every $v \in V$.

\begin{figure}
  \begin{center}
    \begin{tikzpicture}[thick]
      \begin{scope}[scale=0.5]
        \node at (-1,2) {a)};
        \foreach \x in {0,...,5}{
          \foreach \y in {0,...,4}{
            \node [draw, circle, dashed, black!50] at (\x, \y) {};
          }
        }
        \foreach \x in {0,...,4}{
          \foreach \y in {0,...,4}{
            \path[-]
            (\x,\y) edge[dashed, black!50] (\x+1,\y);
          }
        }
        \foreach \x in {0,...,5}{
          \foreach \y in {0,...,3}{
            \path[-]
            (\x,\y) edge[dashed, black!50] (\x,\y + 1);
          }
        }
        \foreach \x in {0,...,5}{
          \foreach \y in {0,...,4}{
            \node(\x\y) [draw = black!50, dashed, circle,fill=white] at (\x, \y) {};
          }
        }
        \node at (3,2){$v_0$};
        \node at (2,2){$v_1$};
        \node at (1,2){$v_2$};
        \node at (1,3){$v_3$};

        \node[draw = black, circle] at (3,2){};
        \node[draw = red, circle] at (2,2){};
        \node[draw = green!50!black, circle] at (4,2){};
        \node[draw = blue, circle] at (3,3){};
        \node[draw = purple, circle] at (3,1){};
      \end{scope}
      \begin{scope}[xshift=5cm, scale=0.5]
        \node at (-1,2) {b)};
        \foreach \x in {0,...,5}{
          \foreach \y in {0,...,4}{
            \node [draw, circle, dashed, black!50] at (\x, \y) {};
          }
        }
        \foreach \x in {0,...,4}{
          \foreach \y in {0,...,4}{
            \path[-]
            (\x,\y) edge[dashed, black!50] (\x+1,\y);
          }
        }
        \foreach \x in {0,...,5}{
          \foreach \y in {0,...,3}{
            \path[-]
            (\x,\y) edge[dashed, black!50] (\x,\y + 1);
          }
        }
        \foreach \x in {0,...,5}{
          \foreach \y in {0,...,4}{
            \node(\x\y) [draw = black!50, dashed, circle,fill=white] at (\x, \y) {};
          }
        }
        \node at (3,2){$v_0$};
        \node at (2,2){$v_1$};
        \node at (1,2){$v_2$};
        \node at (1,3){$v_3$};
        \draw
        (0.5,2) -- (3,4.5) -- (5.5,2) -- (3,-0.5) -- (0.5,2);
        \path[->,>=stealth']
        (31) edge (21)
        (32) edge (22)
        (33) edge (23)
        (42) edge (32)
        (22) edge (12);
        \node[draw = black, circle] at (2,2){};
        \node[draw = red, circle] at (1,2){};
        \node[draw = green!50!black, circle] at (3,2){};
        \node[draw = blue, circle] at (2,3){};
        \node[draw = purple, circle] at (2,1){};
      \end{scope}
      \begin{scope}[scale=0.5, yshift=-6cm]
        \node at (-1,2) {c)};
        \foreach \x in {0,...,5}{
          \foreach \y in {0,...,4}{
            \node [draw, circle, dashed, black!50] at (\x, \y) {};
          }
        }
        \foreach \x in {0,...,4}{
          \foreach \y in {0,...,4}{
            \path[-]
            (\x,\y) edge[dashed, black!50] (\x+1,\y);
          }
        }
        \foreach \x in {0,...,5}{
          \foreach \y in {0,...,3}{
            \path[-]
            (\x,\y) edge[dashed, black!50] (\x,\y + 1);
          }
        }
        \foreach \x in {0,...,5}{
          \foreach \y in {0,...,4}{
            \node(\x\y) [draw = black!50, dashed, circle,fill=white] at (\x, \y) {};
          }
        }
        \node at (3,2){$v_0$};
        \node at (2,2){$v_1$};
        \node at (1,2){$v_2$};
        \node at (1,3){$v_3$};
        \draw
        (-0.5,2) -- (2,4.5) -- (4.5,2) -- (2,-0.5) -- (-0.5,2);
        \path[->,>=stealth']
        (21) edge (11)
        (22) edge (12)
        (23) edge (13)
        (32) edge (22)
        (12) edge (02);
        \node[draw = black, circle] at (1,2){};
        \node[draw = red, circle] at (0,2){};
        \node[draw = green!50!black, circle] at (2,2){};
        \node[draw = blue, circle] at (1,3){};
        \node[draw = purple, circle] at (1,1){};
      \end{scope}
      
      \begin{scope}[xshift=5cm, scale=0.5,yshift=-6cm]
        \node at (-1,2) {d)};
        \foreach \x in {0,...,5}{
          \foreach \y in {0,...,4}{
            \node [draw, circle, dashed, black!50] at (\x, \y) {};
          }
        }
        \foreach \x in {0,...,4}{
          \foreach \y in {0,...,4}{
            \path[-]
            (\x,\y) edge[dashed, black!50] (\x+1,\y);
          }
        }
        \foreach \x in {0,...,5}{
          \foreach \y in {0,...,3}{
            \path[-]
            (\x,\y) edge[dashed, black!50] (\x,\y + 1);
          }
        }
        \foreach \x in {0,...,5}{
          \foreach \y in {0,...,4}{
            \node(\x\y) [draw = black!50, dashed, circle,fill=white] at (\x, \y) {};
          }
        }
        \node at (3,2){$v_0$};
        \node at (2,2){$v_1$};
        \node at (1,2){$v_2$};
        \node at (1,3){$v_3$};
        \draw
        (-1.5,2) -- (1,4.5) -- (3.5,2) -- (1,-0.5) -- (-1.5,2);
        \path[->,>=stealth']
        (11) edge (12)
        (12) edge (13)
        (13) edge (14)
        (22) edge (23)
        (02) edge (03);
        \node[draw = black, circle] at (1,3){};
        \node[draw = red, circle] at (0,3){};
        \node[draw = green!50!black, circle] at (2,3){};
        \node[draw = blue, circle] at (1,4){};
        \node[draw = purple, circle] at (1,2){};
      \end{scope}
    \end{tikzpicture}
  \end{center}
  \caption{Illustration of the translation of a small indexing kernel using translations in each induced subgraph. Kernel is initialized around $v_0$ (a), then moved left around $v_1$ (b) using the induced subgraph around $v_0$, then moved left again around $v_2$ (c) using the induced subgraph around $v_1$ then moved up around $v_3$ (d) using the induced subgraph around $v_2$. At the end of the process, the kernel has been localized around each vertex in the graph.}
  \label{movekernel}
\end{figure}
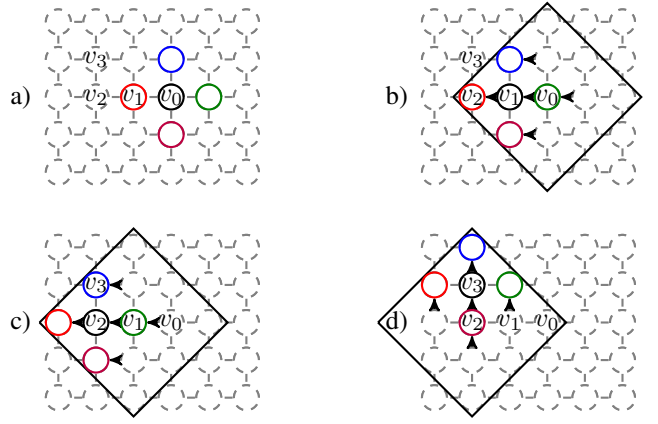

This process is depicted in Figure~\ref{movekernel}. Since it requires moving the kernel everywhere, its complexity is $\mathcal{O}(n d^2)$.\\

\noindent\textbf{Final step: identifying proxy-translations in $G$}

Finally, by looking at the indexings obtained in the previous step, we obtain a family of proxy-translations defined globally on $G$. More precisely, each index defines its own proxy-translation. Note that they are not translations because only the local properties have been propagated through the second step, so there can exist aligned candidates with smaller losses. Because of the constraint to keep the paths with the minimum sum of losses, they are good proxies to translations on $G$.

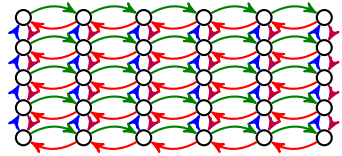
\begin{figure}
  \begin{center}
    \begin{tikzpicture}[thick,scale=0.8]
      \tikzstyle{every node} = [inner sep = 2pt];
      \begin{scope}
        \foreach \x in {0,...,5}{
          \foreach \y in {0,...,4}{
            \node(\x\y) [draw, circle, black] at (\x, \y/2) {};
          }
        }
        \foreach \x in {0,...,4}{
          \foreach \y in {0,...,4}{
            \pgfmathtruncatemacro{\nouveaux}{\x+1}
            \path[->,>=stealth']            
            (\x\y) edge[bend left, green!50!black] (\nouveaux\y);
          }
        }
        \foreach \x in {0,...,5}{
          \foreach \y in {0,...,3}{
            \pgfmathtruncatemacro{\nouveauy}{\y+1}
            \path[->,>=stealth']
            (\x\y) edge[bend left, blue] (\x\nouveauy);
          }
        }
        \foreach \x in {1,...,5}{
          \foreach \y in {0,...,4}{
            \pgfmathtruncatemacro{\nouveaux}{\x-1}
            \path[->,>=stealth']
            (\x\y) edge[bend left, red] (\nouveaux\y);
          }
        }
        \foreach \x in {0,...,5}{
          \foreach \y in {1,...,4}{
            \pgfmathtruncatemacro{\nouveauy}{\y-1}
            \path[->,>=stealth']
            (\x\y) edge[bend left, purple] (\x\nouveauy);
          }
        }
      \end{scope}
    \end{tikzpicture}
  \end{center}
  \caption{Proxy-translations in $G$ obtained after moving the small kernel around each vertex. Each color corresponds to one translation.}
  \label{translations}
\end{figure}

An illustration on a grid graph is given in Figure~\ref{translations}. The complexity is $\mathcal{O}(nd)$. Overall, all three steps are linear in $n$.

\subsection{Extended Convolution Layers}
\label{extconv}

Let $(\psi_p)_{p=0,..,\kappa-1}$ be the proxy-translations identified on $G$ with the convention that $\psi_0 = id$ is the identity function, and where $\kappa$ is the number of weights in the indexing kernel.

The operation of the \emph{extended convolution layer} centered on the vertex $v \in V$ is defined as:

$$\mathbf{y}_v = h\left(\sum_{p=0}^{\kappa-1}{w_p \mathbf{x}_{\phi_p(v)}} + b\right)$$

where $h$ is the activation function, $b$ is the bias term, $\mathbf{x}_\bot = 0$ and:
$$
\left\{
    \begin{array}{ll}
        \phi_p(v) = \psi_p(v) & \mbox{ if } \psi_p \mbox{ is defined on } v \\
        \phi_p(v) = \bot \notin V & \mbox{ else} \\
    \end{array}
\right..
$$

Note that we defined convolution layers using the formalism of proxy-translations, but they can also be defined using only the formalism of local translations~\cite{pasdeloup2017convolutional}.

\subsection{Extended Data Augmentation}

Once translations are obtained on $G$, one can use them to move training vectors, artificially creating new ones. Note that this type of data-augmentation is poorer than for images since no flipping, scaling or rotations are used.

\subsection{Extended Downscaling Layers}

Downscaling is a tricky part of the process because it supposes one can somehow regularly sample vectors. As a matter of fact, a nonregular sampling is likely to produce a highly irregular downscaled graph, on which looking for translations irremediably leads to poor accuracy, as we noticed in our experiments.

We rather define the translations of the strided graph using the previously found proxy-translations on $G$.\\

\noindent\textbf{First step: extended convolution with stride $r$}

Given an arbitrary initial vertex $v_0 \in V$, the set of kept vertices $V_{\downarrow r}$ is defined inductively as follows:
\begin{itemize}[noitemsep,nolistsep]
\item $V_{\downarrow r}^0 = \{v_0\}$,
\item $\forall t \in \mathbb{N}, V_{\downarrow r}^{t+1} = V_{\downarrow r}^t \cup \{v \in V, \forall v' \in V_{\downarrow }^t, v \not\in N_{r-1}(v') \land \exists v' \in V_{\downarrow r}^t, v \in N_{r}(v') \}$.
\end{itemize}

\begin{figure}
  \begin{center}
    \begin{tikzpicture}[thick, scale=0.6]
      \tikzstyle{every node} = [inner sep = 2pt];
      \foreach \x in {0,...,5}{
        \foreach \y in {0,...,4}{
          \node [draw, circle, black!50] at (\x, \y/2) {};
        }
      }
      \foreach \x in {0,...,4}{
        \foreach \y in {0,...,4}{
          \path[-]
          (\x,\y/2) edge[black!50] (\x+1,\y/2);
        }
      }
      \foreach \x in {0,...,5}{
        \foreach \y in {0,...,3}{
          \path[-]
          (\x,\y/2) edge[black!50] (\x,\y/2 + 1/2);
        }
      }
      \foreach \x in {0,...,5}{
        \foreach \y in {0,...,4}{
          \pgfmathtruncatemacro{\sum}{\x+\y}
          \ifodd\sum
          \node(\x\y) [draw = black!50, circle,fill=white] at (\x, \y/2) {};
          \else
          \node(\x\y) [draw = black!50, circle,fill=black] at (\x, \y/2) {};
          \fi
        }
      }
      \node at (3.4,1.2){\textcolor{blue}{$v_0$}};
      \node[fill=blue!20!white] at (3,1){};
    \end{tikzpicture}
  \end{center}
  
  \caption{Downscaling of the grid graph. Disregarded vertices are filled in.}
  \label{downscaling}
\end{figure}
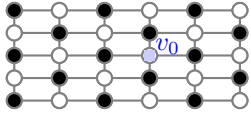

This sequence is nondecreasing and bounded by $V$, so it eventually becomes stationary and we obtain $V_{\downarrow r} = \lim_t{V_{\downarrow r}^t}$. Figure~\ref{downscaling} illustrate the first downscaling $V_{\downarrow 2}$ on a grid graph. 

The output neurons of the extended convolution layer with stride $r$ are $V_{\downarrow r}$.\\

\noindent\textbf{Second step: convolutions for the strided graph}

Using the proxy-translations on $G$, we move a localized $r$-hop indexing kernel over $G$. At each location, we associate the vertices of $V_{\downarrow r}$ with indices of the kernel, thus obtaining what we define as induced $_{\downarrow r}$-translations on the set $V_{\downarrow r}$. In other words, when the kernel is centered on $v_0$, if $v_1 \in V_{\downarrow r}$ is associated with the index $p_0$, we obtain $\phi_{p_0}^{\downarrow r}(v_0) = v_1$. Subsequent convolutions at lower scales are defined using these induced $_{\downarrow r}$-translations similarly to Subsection C.


\section{Experiments}
\label{experiments}

To validate our method we performed experiments with two different datasets, CIFAR-10~\cite{krizhevsky2009learning} and PINES fMRI dataset \cite{chang2015sensitive}. The code is available at \href{https://github.com/brain-bzh/MCNN}{github.com/brain-bzh/MCNN}.

\subsection{CIFAR-10}

On the CIFAR-10 dataset, our models are based on a variant of a deep residual network, namely PreActResNet18\cite{he2016identity}. We tested different combinations of graph support and data augmentation. For the graph support, we use either a regular 2D grid or either an inferred graph obtained by keeping the four neighbours that covary the most. Table \ref{cifar-table} summarizes out results. In particular, it is interesting to note that results obtained without any structure prior (91.07\%) are only 2.7\% away from the baseline using classical CNNs on images (93.80\%). This gap is even smaller (less than 1\%) when using the grid prior. Also, without priors our method significantly outperforms the others. 

\begin{table}
\begin{center}
\caption{CIFAR-10 result comparison table.}
\vspace{-0.4cm}
\label{cifar-table}
\resizebox{\columnwidth}{!}{%
\begin{tabular}{|l|c|c||c|c|c|}
\hline
\multirow{2}{*}{Support}  & \multirow{2}{*}{\begin{tabular}[c]{@{}c@{}}MLP\\ \cite{LinMK15}\end{tabular}}     & \multirow{2}{*}{CNN} & \multicolumn{2}{c|}{Grid Graph (Given)} & Covariance Graph (Inferred)    \\ \cline{4-6}
                         &                                                                               &                      & \cite{defferrard2016convolutional} & Proposed & Proposed         \\ \hline
Full Data Augmentation         & 78.62\% & \textbf{93.80\%}   & 85.13\%          & 93.94\%            & 92.57\%                                 \\ \hline
Data Augmentation - Flip & ------- & 92.73\%              & 84.41\%          & 92.94\%            & 91.29\%                                \\ \hline
Graph Data Augmentation        & ------- & 92.10\%$^a$              & ----          & 92.81\%            & \textbf{91.07\%}$^b$                     \\ \hline
None                           & 69.62\%$^b$ & 87.78\%           & ----          & 88.83\%         & 85.88\%$^b$                                \\ \hline
\end{tabular}
}
\vspace{-0.2cm}
\end{center}
\footnotesize{$^a$ As the CNN does not have a graph support we used the covariance graph as support for the graph data augmentation.\\ $^b$ No priors about the structure}\\
\vspace{-.6cm}
\end{table}

\subsection{PINES fMRI}

The PINES dataset consists of fMRI scans on 182 subjects, during an emotional picture rating task\cite{chang2015sensitive}. We fetched individual first-level statistical maps (beta images) for the minimal and maximal ratings from \url{https://neurovault.org/collections/1964/}, to generate the dataset. Full brain data was masked on the MNI template and resampled to a 16mm cubic grid, in order to reduce dimensionality of the dataset while keeping a regular geometrical structure. Final volumes used for classification contain 369 signals for each subject and rating. 

We used a shallow network. The results on Table~\ref{iaps-table} show that our method was able to improve over CNNs, MLPs and other graph-based extended convolutional neural networks.

\begin{table}[h]
\centering
\caption{PINES fMRI dataset accuracy comparison table.}
\label{iaps-table}
\begin{tabular}{|l||c|c||c|c|}
\hline
\multicolumn{1}{|l||}{Graph} & \multicolumn{2}{c||}{None} & \multicolumn{2}{c|}{Neighborhood Graph}     \\ \hline
Method                      & MLP & CNN (kernel 1x1)                                 & \cite{defferrard2016convolutional} & Proposed                   \\ \hline
Accuracy                    & 82.62\% & 84.30\%                            & 82.80\%                            & \textbf{85.08\%} \\ \hline
\end{tabular}
\vspace{-.4cm}
\end{table}

\section{Conclusion}

We proposed a new methodology that extends classical convolutional neural networks to irregular domains represented by a graph.
The methodology scales linearly well with the order of the graph. Moreover, training can be performed using existing libraries for deep learning.

We performed experiments and showed that our method is able to match performance of classical convolutional neural networks on images without explicit knowledge about the underlying regular 2D structure. It also significantly outperforms existing extended convolutional neural networks alternatives based on graphs.
We also demonstrated the ability of the method to adapt to slightly irregular domains by performing experiments on a neuroimage dataset.

However, the main limitation is that on highly irregular domains, the obtained translations aren't very helpful to design meaningful convolutions, especially if the degree of the graph varies a lot. Hence this requires to add constraints to the graph inferring step to obtain an exploitable graph if it is not.

Future work includes extending to highly irregular domains, which might require to revisit the definitions of translations. 

\section*{Acknowledgements}
\small{
This work was funded in part with the support of Région Bretagne and computations were performed with the use of  Nvidia GPUs, courtesy of Nvidia.
}
\bibliographystyle{IEEEtran}
\bibliography{refs}

\end{document}